\documentclass[sigconf,screen]{acmart}

\usepackage{multirow}

\AtBeginDocument{%
 }

\copyrightyear{2026}
\acmYear{2026}
\setcopyright{othergov}
\acmConference[GoodIT '26]{International Conference on Information Technology for Social Good}{September 02--04, 2026}{Pisa, Italy}
\acmBooktitle{International Conference on Information Technology for Social Good (GoodIT '26), September 02--04, 2026, Pisa, Italy}
\acmDOI{10.1145/3794786.3830759}
\acmISBN{979-8-4007-2483-1/2026/09}

\begin{document}







\title{Verifiable Disaster Storylines and Causal Knowledge Graphs: A Citation-Grounded Pipeline from Heterogeneous Humanitarian Sources}

\author{Ivan Decostanzi}
\email{ivan.decostanzi@isi.it}
\orcid{0009-0000-2459-8834}
\affiliation{%
 \institution{ISI Foundation}
 \city{Turin}
 \country{Italy}}

\author{Michele Ronco}
\email{michele.ronco@ec.europa.eu}
\orcid{0000-0002-2160-2452}
\affiliation{
 \institution{European Commission, Joint Research Centre (JRC)}
 \city{Ispra}
 \country{Italy}
}

\author{Sergio Consoli}
\email{sergio.consoli@ec.europa.eu}
\orcid{0000-0001-7357-5858}
\affiliation{
 \institution{European Commission, Joint Research Centre (JRC)}
 \city{Ispra}
 \country{Italy}
}

 \author{Christina Corbane}
\email{christina.corbane@ec.europa.eu}
\orcid{0000-0002-2670-1302}
\affiliation{
 \institution{European Commission, Joint Research Centre (JRC)}
 \city{Ispra}
 \country{Italy}
}

\author{Lorenzo Bertolini}
\email{lorenzo.bertolini@ec.europa.eu}
\orcid{0000-0002-1709-9372}
\affiliation{
 \institution{European Commission, Joint Research Centre (JRC)}
 \city{Ispra}
 \country{Italy}
}

\author{Indaco Biazzo}
\email{indaco.biazzo@ec.europa.eu}
\orcid{0000-0002-9897-7543}
\affiliation{
 \institution{European Commission, Joint Research Centre (JRC)}
 \city{Ispra}
 \country{Italy}
}

\author{Daria Mihaila}
\email{daria.mihaila@ec.europa.eu}
\orcid{0009-0001-4595-0494}
\affiliation{
 \institution{European Commission, Joint Research Centre (JRC)}
 \city{Ispra}
 \country{Italy}
}

\author{Manuel Garcia-Herranz}
\email{mherranz@unicef.org}
\orcid{0000-0002-4252-4975}
\affiliation{
 \institution{UNICEF}
 \city{New York}
 \country{USA}
}

\author{Felix Schwebel}
\email{fschwebel@unicef.org}
\orcid{0009-0002-0864-9966}
\affiliation{
 \institution{UNICEF}
 \city{New York}
 \country{USA}
}

\author{Yelena Mejova}
\email{yelena.mejova@isi.it}
\orcid{0000-0001-5560-4109}
\affiliation{%
 \institution{ISI Foundation}
 \city{Turin}
 \country{Italy}}

\author{Kyriaki Kalimeri}
\email{kyriaki.kalimeri@isi.it}
\orcid{0000-0001-8068-5916}
\affiliation{%
 \institution{ISI Foundation \& UNICEF}
 \city{Turin}
 \country{Italy}}

\renewcommand{\shortauthors}{Decostanzi et al.}

\begin{abstract}

Effective humanitarian response depends on the rapid synthesis of heterogeneous, high-volume information sources — a task that routinely exceeds human analytical capacity in the critical early hours of a crisis. We present a pipeline that combines structured disaster records from EM-DAT with unstructured documents from ReliefWeb and the European Media Monitor (EMM) to produce source-grounded disaster storylines and causal knowledge graphs supporting situational awareness for responders and analysts. Using Retrieval-Augmented Generation, the pipeline extracts structured storylines — tabular event profiles covering 17 fields, from severity and key drivers to child-sensitive impact indicators — and constructs causal knowledge graphs where each node and edge is enriched with citation-grounded explanatory narratives, enabling full traceability back to primary sources. We evaluate the system on three diverse crisis use cases through a human evaluation involving 9 domain expert and 9 non-expert evaluators. Results confirm high retrieval precision, strong faithfulness of extracted causal relations, and a clear expert preference for citation-grounded components over ungrounded alternatives. The pipeline is designed to scale to the full EM-DAT catalogue, with the goal of publicly releasing a new collection of disaster stories complementing the EM-DAT database.

\end{abstract}




\maketitle

\section{Introduction}\label{sec:intro}

In the immediate aftermath of a disaster, the ability to gather, synthesise, and act upon factual information is the difference between a coordinated response and a chaotic one. Disaster risk management (DRM) frameworks, utilised by entities such as the European Civil Protection and Humanitarian Aid Operations (ECHO), the United Nations Office for the Coordination of Humanitarian Affairs (UNOCHA), and the International Federation of Red Cross and Red Crescent Societies (IFRC), rely heavily on secondary data sources — news reports, situational updates, and field observations — to identify relief priorities and allocate resources effectively. However, modern disaster response faces an information paradox: while data availability has increased exponentially, the high volume and velocity of unstructured textual information routinely exceed human analytical capacity~\cite{imran_social_media_2020}. Manually reviewing thousands of articles and reports to identify specific causal links — such as the relationship between a flood event and subsequent displacement or disruption of health services — is frequently intractable within the necessary operational timeframes. Standard disaster databases such as EM-DAT\footnote{\url{https://www.emdat.be/}}, while globally recognised, record aggregate impact statistics and systematically exclude smaller-scale events that do not meet reporting thresholds~\cite{pedra_headlines_2026}, leaving significant gaps in situational awareness.

Complementary textual sources — news articles, field assessments, and humanitarian coordination documents — can fill these gaps by capturing qualitative and quantitative details on impacts, exposed populations, local vulnerabilities, and cascading effects that are often absent from conventional event catalogues~\cite{ronco_disaster_2026}. When paired with Retrieval-Augmented Generation (RAG)~\cite{rag}, these unstructured sources can be transformed into coherent, factually anchored narratives. Knowledge graphs (KGs), which represent entities and their relationships as machine-readable triples, offer a particularly effective formalism for organising the resulting information and enabling structured reasoning over it~\cite{chen_ekell_2024, yao_knowledge_2025, tarraga_causal_2024}. Recent work has demonstrated the viability of this approach across diverse hazard types: LLM-driven KG pipelines have been applied to earthquake emergency management~\cite{yao_knowledge_2025}, typhoon tracking~\cite{huang_typhoon_2026}, compound urban crises~\cite{hao_compound_2025}, and flood impact reporting~\cite{colverd_floodbrain_2023, pedra_headlines_2026}. A convergent finding across these systems is that grounding LLM outputs in structured or retrieved knowledge — rather than relying on parametric memory alone — is essential for factual accuracy in high-stakes operational contexts~\cite{chen_ekell_2024, yao_knowledge_2025, hao_compound_2025}. Broader surveys confirm both the momentum and the remaining challenges of deploying LLMs responsibly in humanitarian settings, stressing the need for human oversight, transparency, and explainability~\cite{xu_llm_disaster_2025, lei-etal-2025-harnessing, shahi_governing_2026}.

However, a persistent limitation of existing LLM-driven KG systems is that both generated narratives and graph elements are typically produced without explicit links to the source evidence from which they are derived. In high-stakes operational contexts — where every claim must be verifiable — this lack of traceability undermines user trust and limits practitioner adoption. Moreover, standard databases under-represent dimensions of particular concern to humanitarian actors, including child-specific vulnerabilities such as displacement, casualties, and loss of access to education and health services~\cite{Kadir2025Child_review, hasbiniDatabaseDisasterImpacts2026a}.

Our prior work ~\cite{ronco_disaster_2026} introduced a pipeline that constructs causal KGs from over 3{,}000 global disaster events by combining EM-DAT records with news articles from the European Media Monitor (EMM) \footnote{\url{https://knowledge4policy.ec.europa.eu/europe-media-monitor-emm\_en}} through RAG-based extraction. Concretely, given an EM-DAT event record — e.g., \textit{Flood, Pakistan, June 2023} — the system retrieves relevant news articles and synthesises them into a structured event profile, which we term a \textit{storyline}: a fixed-schema tabular summary capturing dimensions such as severity, key drivers, and impacts on critical services. From this storyline, a causal knowledge graph is extracted, encoding the event's dynamics as subject–predicate–object triples constrained to causes and prevents relations. The present paper extends this framework in several directions:

\begin{itemize}
    \item \textit{Humanitarian reports integration and enriched evidence base:} We incorporate ReliefWeb\footnote{\url{https://reliefweb.int/}} as a complementary source, linking EM-DAT events to ReliefWeb disaster records via GLIDE identifiers~\cite{glidenumber} and merging humanitarian field reports with news-derived documents. To our knowledge, this is the first pipeline to combine these two sources within an LLM-driven causal KG workflow.
    \item \textit{Full source traceability} We introduce a Multi-Shot RAG strategy in which each storyline field is extracted through an independent retrieval-generation cycle, and a secondary validation step that generates explanatory text for every KG node and edge. Both mechanisms ground every output element in explicit citations to the underlying source documents, providing a verifiable audit trail that directly addresses concerns about transparency and explainability in automated DRM pipelines \cite{shahi_governing_2026}. 



    \item \textit{Child-sensitive impact dimensions:} The storyline extraction
  schema is expanded to cover child-specific indicators, including
  displacement, casualties, and loss of access to education and
  health services.

    \item \textit{Exploration platform:} We provide an interactive dashboard enabling exploration and analysis of the enriched causal knowledge graphs, enhanced source-grounded storylines, and child-specific risk indicators. Additionally, users can interact with the underlying database through natural language queries.

    
    \item \textit{Rigorous multi-level evaluation:} We conduct rigorous human
evaluation --- involving 18~independent annotators --- across three diverse
crisis use cases spanning public health, natural disaster, and armed
conflict, including a citation-level assessment of precision and recall of
the generated attributions.
    
\end{itemize}

All source code is publicly available,\footnote{\url{https://github.com/idecost/StoryLine_KG}} and an interactive dashboard for exploring storylines, knowledge graphs, and child-specific risk indicators is accessible at \url{https://idecost.github.io/StoryLine_KG/Viewer}. The remainder of this paper is organised as follows: Section~\ref{sec:methods} details the pipeline, Section~\ref{sec:evaluation} the evaluation protocol, Section~\ref{sec:results} the results, and Section~\ref{sec:conclusions} discusses limitations and future directions.

\section{Methods}
\label{sec:methods}

\begin{figure*}[!thb]
    \centering
    \begin{minipage}{0.75\textwidth}
        \centering
        \includegraphics[width=\textwidth]{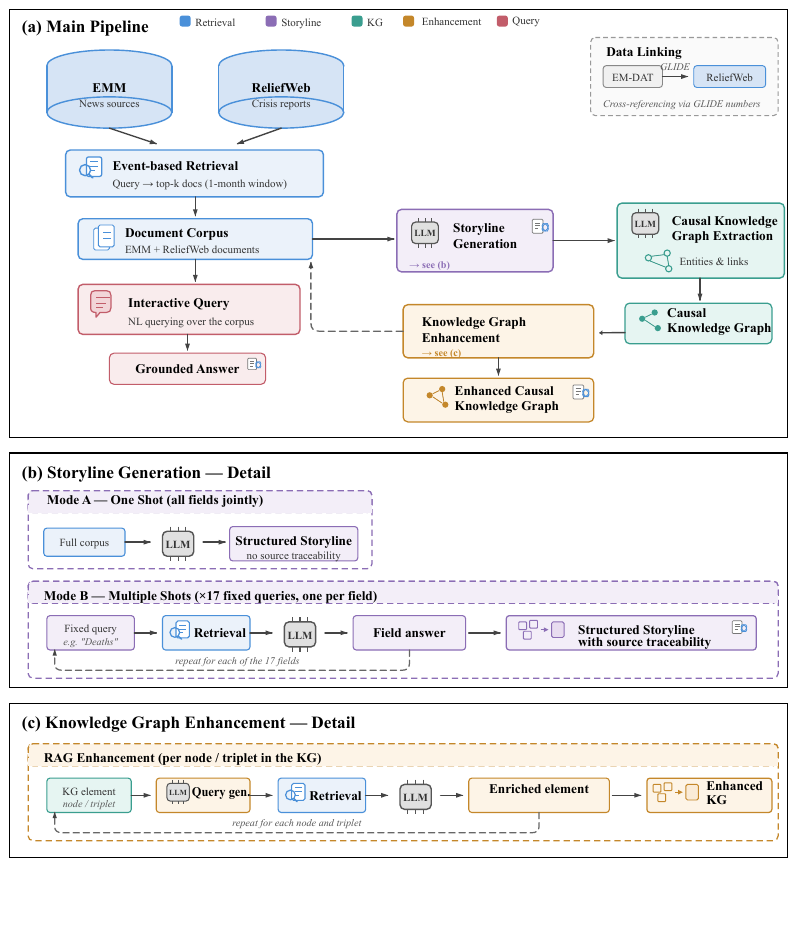}
        \vspace{-60pt} 
        \Description{Flowchart of the pipeline showing data flow from EM-DAT, 
EMM, and ReliefWeb through RAG-based retrieval, storyline generation, 
and knowledge graph construction.}
        \caption{End-to-end pipeline for source-grounded disaster storyline and KG generation. (a) Main pipeline integrating EM-DAT, EMM, and ReliefWeb through RAG-based retrieval. (b) One-Shot vs. Multi-Shot storyline generation strategies. (c) Per-element RAG enhancement of the causal knowledge graph with citation-grounded narratives.}
        \label{fig:pipeline}
    \end{minipage}
\end{figure*}

This section details each component of the pipeline illustrated in Figure~\ref{fig:pipeline}a, highlighting the extensions introduced relative to~\cite{ronco_disaster_2026}. We begin with the integration of ReliefWeb as a complementary humanitarian data source (Section~\ref{sec:reliefweb}), then describe the expanded storyline extraction schema and the two generation strategies (Section~\ref{sec:storyline}), the causal KG construction (Section~\ref{sec:kg_construction}), the citation-grounded validation layer (Section~\ref{sec:factualityKG}), and the natural language query interface (Section~\ref{sec:nl_interface}).

\subsection{ReliefWeb Integration}
\label{sec:reliefweb}

To enrich the information available for each disaster event, we extend
the document retrieval pipeline 
by incorporating humanitarian reports from 
ReliefWeb. This
integration is motivated by the complementary nature of ReliefWeb's
content, which aggregates situation reports, assessments, and
humanitarian coordination documents. These sources often capture
operational details and granular field data absent from news-based
channels, thereby providing a more comprehensive operational picture of
each disaster.

Consistent with the RAG paradigm, where model performance is enhanced by surfacing relevant
external context, we align ReliefWeb data with our existing event-based
structure. To link events across the two databases, we rely on GLIDE
numbers \cite{glidenumber}, a standardized disaster identification
system jointly developed by ADRC, CRED, OCHA/ReliefWeb, and UNDRR. This
system assigns a unique structured code to each disaster, enabling
unambiguous cross-referencing across humanitarian data systems. Each
EM-DAT event is matched to its corresponding ReliefWeb disaster record
using this identifier.

The full set of extracted text is then embedded using the BAAI/bge-m3 model
\cite{bgem3}, with a chunking strategy of four sentences per chunk and a
one-sentence overlap. Candidate chunks are ranked by relevance using
\textit{BAAI/bge-reranker-v2-m3} \cite{bgem3}, and the top~15 chunks are
retained. These ReliefWeb-derived documents are merged with the document
set retrieved from EMM, 
collectively forming the updated evidence base for all downstream tasks.

\subsection{Storyline Generation}
\label{sec:storyline}
 
As illustrated in Figure~\ref{fig:pipeline}a, the first analytical
stage of the pipeline transforms the document corpus into a
structured \textit{storyline}: a set of 17~fields that capture the
key dimensions of a disaster event. These fields span standard impact
indicators --- damage analytics, event mapping, vital service continuity, and contextual threat assessments --- as defined by
established humanitarian frameworks \cite{sendai_framework,
hasbiniDatabaseDisasterImpacts2026a}. Compared to
\cite{ronco_disaster_2026}, we expand the extraction schema to
include child-specific indicators covering casualties, displacement,
and loss of access to education and health services, a dimension that
remains systematically underreported in standard disaster databases
\cite{Kadir2025Child_review}. The complete list of fields is provided
in Table~\ref{tab:storyline_elements}. Not all fields are necessarily
populated for every event, as their availability depends on the
completeness and detail of the underlying source documents.  If no information is available for a given field, it is left as 'Unknown'.


\begin{table}[!b]
\footnotesize
\centering
\caption{Storyline extraction elements grouped by category.}
\label{tab:storyline_elements}
\vspace{-10pt}
\begin{tabular}{p{2.2cm}p{5.4cm}}
\toprule
\textbf{Category} & \textbf{Elements} \\
\midrule
Hazard Profile \& Risk Assessment
  & Key information; Severity; Key drivers;
    Main impacts, exposure, and vulnerability;
    Likelihood of multi-hazard risks \\
Temporal \& Situational Context
  & Temporal details; Phase classification; Non-events \\
Children \& Education Impact
  & Impact on children; Impact on schools \\
Critical Services Disruption
  & Health facilities disrupted;
    Water and sanitation access disrupted \\
Risk Governance \& Best Practices
  & Best practices for managing this risk \\
Recovery \& Response
  & Recommendations and supportive measures for recovery \\
Source Assessment
  & Source type; Confidence level of information;
    Potential reporting bias \\
\bottomrule
\end{tabular}
\end{table}
We implement two alternative strategies for generating storylines,
detailed in Figure~\ref{fig:pipeline}b. They share the same
extraction schema and the same underlying model
(Meta-Llama-3-70B-Instruct \cite{grattafiori2024llama3}), but differ
in how the document corpus is presented to the model and, crucially,
in whether the outputs are traceable to their source evidence.

\subsubsection{One-Shot Approach}
\label{sec:oneshot}
 
The \textit{One-shot} approach serves as the baseline in our evaluation. All documents retrieved for a given event are provided to
the model as a single context block, and the 17~storyline fields are
extracted jointly in one generation step
(Figure~\ref{fig:pipeline}b, Mode~A). This strategy is
straightforward and efficient, as it requires a single model
invocation per event. 
However, because all fields are produced in a single pass over the full corpus, there is no mechanism to trace individual outputs back to their supporting passages.

\subsubsection{Multi-Shot Approach (RAG)}
\label{sec:multishot}
 
To address the traceability limitation of the \textit{One-shot}
strategy, we introduce the \textit{Multi-shot} approach
(Figure~\ref{fig:pipeline}b, Mode~B). Rather than extracting all
fields at once, each of the 17~storyline fields is treated as an
independent information need. For each field, a fixed natural
language query is formulated.  This query is
used to retrieve the most relevant passages from the unified document
corpus via the RAG infrastructure (BGE-M3 embeddings and
BGE-Reranker \cite{bgem3}). The model then generates the field value based on the
retrieved passages, and the supporting source documents are recorded
alongside the output.
 
This decomposition yields two advantages. First, by narrowing the
retrieval scope to a single information need at a time, the model
receives more focused context for each field, reducing the risk that
relevant details are overlooked or conflated within long input
sequences. Second, each output element is naturally accompanied by
explicit references to the source passages that support it, enabling
full traceability from the structured storyline back to the original
EMM and ReliefWeb documents. The 17~individually grounded fields are
then assembled into a complete structured storyline with source
attribution.
 
A comparative evaluation of the \textit{One-shot} and
\textit{Multi-shot} approaches is presented in
Section~\ref{sec:evaluation}, where human annotators assess both
factual accuracy and the perceived value of source attribution.

\subsection{Causal Knowledge Graph Construction}
\label{sec:kg_construction}
The generated storyline, produced by either the \textit{One-shot} or
the \textit{Multi-shot} approach, serves as input for the construction
of a causal knowledge graph, following the text-to-graph
methodology in \cite{ronco_disaster_2026}
(Figure~\ref{fig:pipeline}a, Causal Knowledge Graph Extraction).
By construction, graph elements --- nodes and triplets --- are compact
abstractions detached from the textual evidence from which they
originate. While their accuracy can be assessed through manual
inspection against the source documents, as demonstrated in the
evaluation of \cite{ronco_disaster_2026} and in
Section~\ref{sec:results}, this process is labour-intensive and
does not scale to the thousands of elements produced across a large
event catalogue. To address this, we introduce an automated
citation-grounded validation layer (Section~\ref{sec:factualityKG})
that enriches every KG element with explanatory text and explicit
source references, enabling practitioners to verify the factual
basis of each node and edge without revisiting the full document
corpus.

\subsection{Citation-Grounded KG Validation}
\label{sec:factualityKG}
Our validation approach follows the ``attribute-then-generate''
paradigm \cite{attributeGenerateRAG, qian2024capacityCitations},
adapted to the KG setting drawing on LLM-based fact verification
methods \cite{xue_knowledge_2022, shami_fact_2025, huaman_knowledge_2020}.
As illustrated in Figure~\ref{fig:pipeline}c, the framework operates
as a secondary RAG-based pipeline applied independently to every
node and triplet in the KG. For each element, the process proceeds
in three steps:

    (1) \textit{Query formulation.} The LLM converts the KG element
    --- a node label or a subject--predicate--object triplet --- into
    a natural language question designed to retrieve supporting
    evidence. Unlike the \textit{Multi-shot} storyline generation,
    where queries are fixed and predefined, here they are generated
    dynamically, since KG elements emerge from the extraction process
    and cannot be anticipated in advance (e.g., a node labelled
    \textit{``infrastructure damage''} might yield the query
    \textit{``What infrastructure was damaged during the event?''}).

    (2) \textit{Grounded retrieval.} The generated query is executed
    against the unified embedding space containing both ReliefWeb and
    EMM documents, using the same BGE-M3 and BGE-Reranker models
    employed throughout the pipeline.

    (3) \textit{Narrative synthesis.} The retrieved passages are
    used to generate a concise explanatory narrative that
    contextualises the graph element within the broader disaster
    account, rather than leaving it as an isolated structural
    statement.

\begin{figure*}[!thb]
    \centering
    \begin{minipage}{0.87\textwidth}
        \centering
        \includegraphics[width=\textwidth]{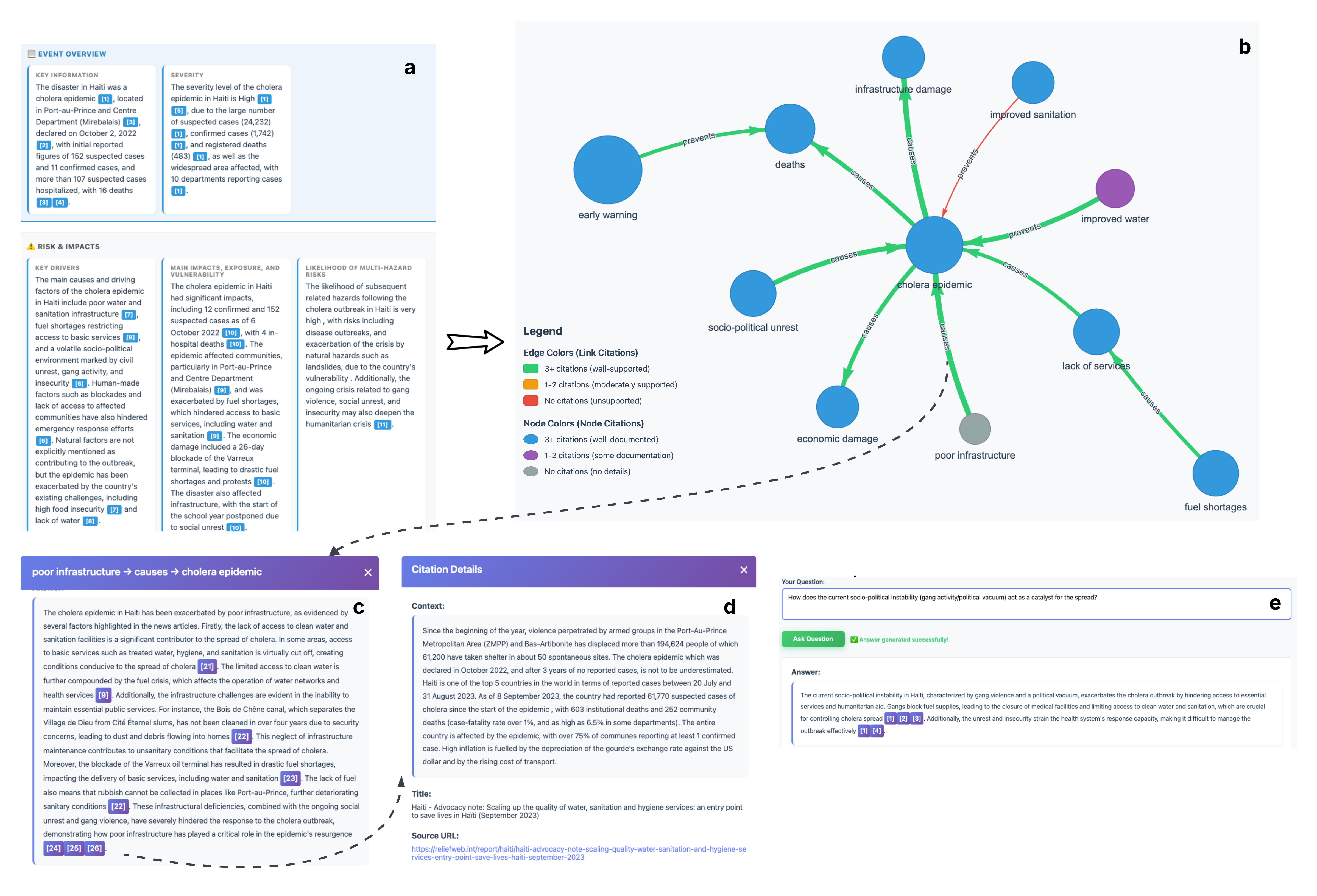}
        \vspace{-30pt}
        \Description{Screenshot of the interactive dashboard showing a 
storyline excerpt, causal knowledge graph, source-grounded narrative, 
citation popup, and natural language query interface.}
        \caption{Example output for the Haiti cholera case study. (a) Excerpt of the generated storyline with inline citations. (b) Causal knowledge graph derived from the storyline. (c) Source-grounded narrative associated with a selected edge. (d) Citation detail popup showing the retrieved source passage. (e) RAG-based Q\&A answering a user query beyond the scope of the storyline and knowledge graph.}
        \label{fig:dashboard}
    \end{minipage}
\end{figure*}

Each generated narrative is accompanied by explicit citations to the
source documents and their metadata, producing the \textit{Enhanced Causal
Knowledge Graph} shown in Figure~\ref{fig:pipeline}a. This audit trail
allows practitioners to verify whether each node and relationship is
factually supported by official sources, directly mitigating the
propagation of hallucinated content through the knowledge graph.
An example of the output is presented in Figure \ref{fig:dashboard}(b,c,d).
\subsection{Natural Language Database Interaction}
\label{sec:nl_interface}
 
The structured outputs described above --- storylines and knowledge
graphs --- capture the information the pipeline is designed to
extract. However, users may have questions that fall outside the
scope of the predefined 17~fields or the graph structure, such as
cross-event comparisons or queries about contextual factors not
covered by the extraction schema. To support such exploratory
analysis, we provide a natural language query interface to the
underlying database. Users can formulate free-text questions about disaster events, impacts, and contextual factors. Each question is translated into a
structured database query by the LLM, executed against the stored
records, and the resulting answer is supported by retrieved textual
evidence when available, ensuring that responses remain grounded in
the source material. This functionality enables non-technical
stakeholders --- including field coordinators and policy analysts ---
to discover event-level details and cross-event patterns that may not
surface through the structured pipeline outputs alone.

\section{Evaluation}
\label{sec:evaluation}

This section presents the evaluation of the pipeline across multiple
dimensions. 
We describe the three crisis use cases selected for
assessment (Section~\ref{sec:usecases}), as well as the employed human
evaluation protocol (Section~\ref{sec:human_eval}). The pipeline outputs — including the causal knowledge graphs, source-grounded storylines, and 
risk indicators — are publicly accessible through an interactive exploration dashboard at \url{https://idecost.github.io/StoryLine_KG/Viewer/}, which showcases a set of precomputed humanitarian events alongside the three use cases employed in this evaluation.

\subsection{Use Cases}
\label{sec:usecases}


We evaluate the pipeline on three events spanning distinct humanitarian
crisis typologies: a public health emergency, a sudden-onset natural
disaster, and a protracted conflict. The cases were selected for their
humanitarian significance, the richness of their documentation across
EM-DAT and ReliefWeb, and their diversity along dimensions relevant to
pipeline performance --- geographic context, temporal dynamics, and
reporting density.

\textbf{Event~1 --- Haiti Cholera Outbreak (2022).}
The resurgence of cholera in Haiti in October 2022 occurred against a
backdrop of political instability, gang violence disrupting aid
delivery, and collapsed water and sanitation infrastructure\footnote{\url{https://www.cdc.gov/mmwr/volumes/72/wr/mm7202a1.htm}}. The
outbreak spread to all ten departments, resulting in tens of thousands
of suspected cases, with children under five disproportionately
affected. Its multi-causal nature --- intersecting public health,
governance failure, and conflict --- makes it a demanding test of the
pipeline's ability to extract complex causal chains.

\textbf{Event~2 ---Hurricane Melissa, Dominican Republic (2025).}
Hurricane Melissa\footnote{\url{https://en.wikipedia.org/wiki/Hurricane_Melissa}} caused widespread displacement and infrastructure
damage across the Caribbean in late 2025. We focus on the Dominican
Republic, where the storm disrupted roads, schools, and health
facilities across southern and eastern regions. The event is
well-documented through OCHA and Dominican Civil Defence situation
reports on ReliefWeb, providing a strong reference for evaluating KG
faithfulness. Its rapid-onset, geographically bounded character offers
a useful contrast to the other two cases.

\textbf{Event~3 --- Syria Conflict Escalation (Late 2024).}
The armed conflict escalation beginning in November 2024 ---
culminating in the fall of Aleppo and the collapse of the Assad
government --- triggered the displacement of over one million people
within weeks, alongside widespread destruction of civilian
infrastructure and acute food insecurity\footnote{\url{https://www.hrw.org/world-report/2024/country-chapters/syria}}. As the most
information-dense and structurally complex of the three cases, it
tests the pipeline's capacity to handle rapidly evolving conflict
settings with fragmented and heterogeneous source material.





\subsection{Human Evaluation}
\label{sec:human_eval}

We conduct an extensive human evaluation of
the full pipeline across the three use cases. The evaluation is carried
out by 18~independent annotators --- 9~domain experts and 9~non-experts
--- none of whom are affiliated with the project or have conflicts of
interest. 
Each evaluated item has been judged by nine different annotators. To assess the
reliability of the collected annotations, we compute multiple
complementary agreement measures. As the primary measure, we report Krippendorff's
$\alpha$~\cite{krippendorff}, a standard metric for multi-annotator
settings that accommodates ordinal scales and missing data. We
additionally compute Fleiss' $\kappa$ as a cross-check; consistent with
the literature~\cite{kappaAlpha}, we find that Fleiss' $\kappa$ yields
values very close to Krippendorff's $\alpha$ across all tasks, and
therefore do not report it separately. We also compute, for each pair of annotators,
the raw percentage agreement and the Prevalence-Adjusted Bias-Adjusted
Kappa (PABAK)~\cite{pabak}, a variant of Cohen's $\kappa$ that
corrects for prevalence and response-bias artefacts, and report the mean values across all annotator pairs. In summary, we report Krippendorff's $\alpha$, mean pairwise percentage agreement, and mean pairwise PABAK as our agreement measures.

Different evaluation stages, detailed in the following, are assigned to the most
appropriate annotator profile: non-experts assess retrieval quality, 
KG text quality, and citation quality as these tasks require no domain-specific knowledge;
experts evaluate storyline quality, KG faithfulness, and the overall
system through the interactive dashboard.

\subsubsection*{Retrieval Quality.}
\label{sec:eval_retrieval}

While automatic RAG evaluation frameworks
exist~\cite{esRAGAsAutomatedEvaluation2024,
salemiEvaluatingRetrievalQuality2024}, their application to the
humanitarian domain remains largely unexplored. We therefore perform a
precision-oriented human assessment: for each retrieved chunk (from
ReliefWeb or EMM), non-expert annotators judge
(a)~\emph{relevance} --- whether the paragraph refers to the specific
target disaster, mentioning the correct location, timeframe, or hazard
type --- and (b)~\emph{informativeness} --- whether it contains
concrete facts or analysis, as opposed to boilerplate, fragments, or
garbled text.

\subsubsection*{Storyline Quality.}
\label{sec:eval_storyline}

Domain experts compare the two storyline generation approaches --- One-shot and Multi-shot --- on a field-by-field basis across the 17 categories in Table~\ref{tab:storyline_elements}. For each field, evaluators select one of the options: One-shot is better, Multi-shot is better, or Cannot tell. Additionally, evaluators provide an overall quality rating for each storyline on a 1--5 Likert scale. This design enables fine-grained analysis of which information categories benefit from each approach, as well as a holistic assessment of the two generation strategies.

\subsubsection*{Knowledge Graph Text Quality.}
\label{sec:eval_kg_text}

For the explanatory texts generated for each KG node and link
(Section~\ref{sec:factualityKG}), non-expert annotators evaluate two
independent dimensions. \emph{Relevance}: whether the generated text
discusses the correct concept (for nodes) or the correct relationship
between source and target (for links), as opposed to being off-topic or
addressing unrelated concepts. \emph{Informativeness}: whether the text
provides concrete, useful information about the disaster event, rated
on a three-point scale --- \emph{very informative} (rich in facts,
figures, or details), \emph{quite informative} (some useful content but
limited in depth), or \emph{not informative} (empty, boilerplate, or
tautological). Nodes and links are assessed in separate sub-tasks.

\subsubsection*{Knowledge Graph Faithfulness.}
\label{sec:eval_kg_faith}

Following~\cite{ronco_disaster_2026}, domain experts evaluate whether
the extracted triples are supported by the storyline from which they are
derived. Each triple
(\textit{Source}~$\rightarrow$~\textit{Relation}~$\rightarrow$~\textit{Target})
is classified as: \emph{Fully supported} (explicitly stated),
\emph{Partially supported} (one entity or an implicit relation is
present), \emph{Not present}, or \emph{Cannot determine}.

\subsubsection*{Citation Quality.}
\label{sec:eval_citation}

We additionally assess the citations attached to storyline fields
(Section~\ref{sec:multishot}) and to KG node and link narratives
(Section~\ref{sec:factualityKG}), adopting the \emph{citation recall} and
\emph{citation precision} definitions established in prior
work~\cite{gao2023-citations, decostanzi2025large}.
Given a generated text composed of statements $s_1,\ldots,s_n$, where each
$s_i$ cites a set of passages $C_i=\{c_{i,1},c_{i,2},\ldots\}$, citation
recall evaluates whether $s_i$ is supported by $C_i$ as a whole, while
citation precision evaluates whether each individual citation $c_{i,j}$ is
relevant to $s_i$; the two coincide when a statement carries a single
citation. Non-expert annotators judge both dimensions on a three-point
scale --- \emph{fully supports}, \emph{partially supports}, \emph{does not
support} --- applied to the individual citation (precision) or to the
citation set (recall).

\subsubsection*{Expert System Assessment.}
\label{sec:eval_system}

Finally, domain experts interact with the full exploration dashboard
--- including the knowledge graph visualisation, source-grounded
storylines, and the natural language database interface
(Section~\ref{sec:nl_interface}) --- and provide a holistic evaluation
of the system. Following established usability assessment
practices~\cite{usabilityScale, ronco_disaster_2026}, experts respond
to structured questions covering the perceived usefulness of individual
components, the utility of the system for humanitarian workflows, and
overall trust in the generated outputs. Responses are collected on a
five-point Likert scale, except for overall trust which is rated on a 0--10 scale.

\section{Evaluation Results}
\label{sec:results}

All agreement metrics are computed pooling annotations across the three 
events; per-event breakdowns are available upon request. Each task is assessed 
by all 9~annotators of the relevant profile.

A recurring pattern across tasks is that Krippendorff's $\alpha$ remains 
modest, while mean pairwise percentage agreement and PABAK are considerably 
higher. This is expected given the strong label imbalance present in most 
tasks---for instance, the large majority of retrieved paragraphs are judged 
relevant, and most KG texts are rated as relevant to their concept. As 
discussed in Section~\ref{sec:human_eval}, 
both $\kappa$- and $\alpha$-family 
statistics are known to be deflated under such prevalence conditions~\cite{Marzi2024reliability}, 
making PABAK and raw percentage agreement more informative indicators of 
true annotator consistency in this setting.

\subsubsection*{Retrieval Quality}

A total of 110~paragraphs are evaluated across the three events. For 
\emph{relevance}, annotators reach 83.1\% mean pairwise agreement 
(PABAK~$= 0.662$, Krippendorff's $\alpha = 0.305$, fair agreement~\cite{KrippendorffContentAA}). 
On average, $85.8 \pm 4.7$\% of retrieved paragraphs are judged relevant 
to the target event, confirming the high precision of the retrieval stage. 
For \emph{informativeness}, agreement is somewhat lower (67.2\% mean 
pairwise, PABAK~$= 0.344$, $\alpha = 0.187$, fair), reflecting the greater 
subjectivity of distinguishing substantive content from boilerplate; 
$72.0 \pm 15.1$\% of paragraphs are nonetheless deemed meaningful.

\subsubsection*{Storyline Quality}
\label{sec:storyline_quality}

A total of 51~field comparisons are evaluated across the three events
(17~fields per event). For each field, evaluators choose \emph{Multi-shot is better},
\emph{One-shot is better}, or \emph{Cannot tell}; they additionally rate
each storyline as a whole on a 1--5 scale. Agreement on this task is substantially lower than on retrieval
and KG quality: pooled across all three events, annotators reach only
51.6\% mean pairwise agreement (PABAK~$= 0.033$, Krippendorff's
$\alpha = 0.097$, slight agreement), reflecting the inherent
subjectivity of comparing narrative outputs.

Results vary considerably across events. For Event~1, the Multi-shot
storyline is preferred in $62.7 \pm 9.0$\% of fields, One-shot in
$19.6 \pm 6.8$\%, and $17.6 \pm 5.9$\% are judged \emph{Cannot tell}.
Event~3 shows a similar pattern with a stronger preference for
Multi-shot ($68.6 \pm 9.0$\% vs.\ $23.5 \pm 17.6$\%). Event~2 diverges
markedly, driven by sharp annotator disagreement---one expert preferred
Multi-shot in 94.1\% of fields while another preferred One-shot in
58.8\%. Overall, the Multi-shot storyline is preferred in
$62.1 \pm 20.0$\% of fields, One-shot in $26.1 \pm 19.5$\%, and
$11.8 \pm 9.3$\% are judged \emph{Cannot tell}.

Overall quality ratings confirm a mild preference for the Multi-shot
approach. Multi-shot receives a mean rating of $3.67 \pm 0.71$ out of~5
across all nine annotators, compared to $2.78 \pm 0.83$ for One-shot,
and seven out of nine annotators express an overall preference for
Multi-shot. Annotators consistently value the source citations present
in Multi-shot storylines, deemed essential for verifiability, though
some fields are better served by the more concise One-shot output.

\subsubsection*{Knowledge Graph Text Quality}
A total of 35~node texts and 28~link texts are evaluated. For KG \emph{node}
texts, $94.0 \pm 2.7$\% of node texts are judged relevant to their concept
(mean pairwise agreement 90.8\%, PABAK~$= 0.816$, $\alpha = 0.190$). The
low $\alpha$ relative to the high percentage agreement is consistent with
the near-ceiling label distribution. Informativeness agreement is more
moderate (60.0\% mean pairwise, PABAK~$= 0.200$, $\alpha = 0.302$, fair);
$89.6 \pm 6.3$\% of node texts are rated at least quite
informative---$35.6 \pm 13.2$\% very informative and $54.0 \pm 9.0$\%
quite informative---while only $10.5 \pm 8.1$\% are rated not informative.
The high standard deviations on the two positive categories suggest that
disagreement concentrates on the boundary between very and quite informative
rather than on whether a text is informative at all.

For KG \emph{link} texts, relevance follows a similar pattern
($92.3 \pm 4.9$\% judged relevant; mean pairwise agreement 88.6\%,
PABAK~$= 0.771$, $\alpha = 0.201$, fair). Informativeness agreement is
lower (55.5\% mean pairwise, PABAK~$= 0.109$, $\alpha = 0.174$, slight),
reflecting the inherent difficulty of assessing the informational depth of
short relational descriptions; $94.4 \pm 9.5$\% of link texts are rated at
least quite informative---$54.4 \pm 14.1$\% very informative and
$40.1 \pm 14.6$\% quite informative---while only $5.6 \pm 5.1$\% are rated
not informative. As with node texts, the high variability across annotators
on the two positive categories confirms that the distinction between very
and quite informative is inherently subjective.

\subsubsection*{Knowledge Graph Faithfulness}
A total of 28~triples are assessed across the three events, with no
substantial variation observed across individual disasters. Overall,
$86.7\%$ of triples are judged as supported by the
storyline---$56.1 \pm 26.5$\% fully supported and
$30.6 \pm 20.7$\% partially supported---while only $9.2 \pm 9.4$\% are
rated as \emph{not present} and $4.2 \pm 12.5$\% as \emph{cannot
determine}. These results confirm that the large majority of automatically
extracted causal relations are grounded in the generated narrative. Inter-annotator agreement is modest (Krippendorff's $\alpha = 0.222$, mean
pairwise agreement $= 54.2\%$, PABAK $= 0.083$). As in previous tasks, disagreement concentrates on the boundary between the two positive categories rather than on whether a triple is supported at all.

\subsubsection*{Citation Quality}
\label{sec:citation_quality}
A total of 391~unique citations and 261~claims are assessed across 70~items and the
three events. Pooled over all components, $91.7$\% of citations are at least partially
relevant to the statement they support ($79.3$\% relevant, $12.4$\% partially relevant;
Krippendorff's $\alpha = 0.515$, mean pairwise agreement $= 77.4$\%, PABAK~$= 0.662$),
and $93.5$\% of claims are at least partially supported by their citation set
($87.1$\% supported, $6.4$\% partially supported; $\alpha = 0.465$, mean pairwise
agreement $= 86.4$\%, PABAK~$= 0.796$). Performance is uneven across components: KG node
and link citations are consistently reliable ($95.7$\% and $92.6$\% at least partially
relevant; $99.1$\% and $98.4$\% of claims at least partially supported), whereas
storyline citations show substantially more variability ($71.7$\% and $65.4$\%,
respectively). This appears to stem from over-citation: in the storyline the model
attaches citations even when it lacks the evidence to answer, so short fields often
carry several irrelevant sources. This reluctance to abstain deserves dedicated
investigation, as citation errors disproportionately affect trust in this domain.

\subsubsection*{Expert System Assessment}

Nine domain experts interact with the exploration dashboard and, as described in the following, they provide holistic evaluations across three dimensions: \textit{feature utility}, \textit{operational impact}, and \textit{overall trust}.

\begin{figure}[!thb]
    \centering
    \includegraphics[width=0.88\linewidth]{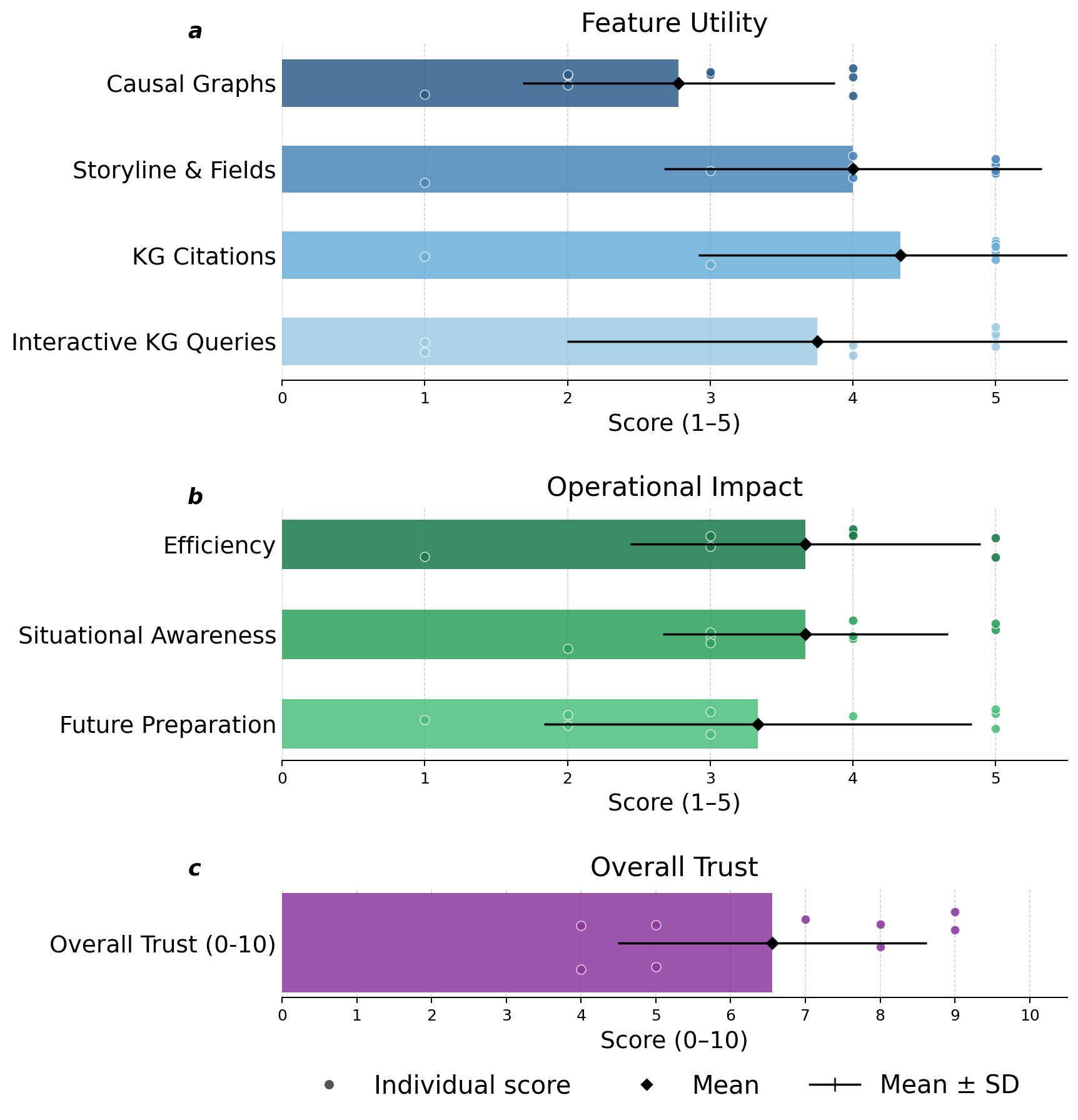}
    \vspace{-10pt} 
    \Description{Bar charts showing expert evaluation scores for feature 
utility, operational impact, and overall trust.}
    \caption{Expert evaluation results from 9 domain experts. (a) Feature utility scores (1–5 scale) for the four main components of the dashboard: causal graphs, storyline fields, knowledge graph citations, and interactive KG queries. (b) Perceived operational impact (1–5 scale) across 3 dimensions: efficiency in crisis analysis workflows, enhancement of situational awareness, and usefulness for future disaster preparation. (c) Overall trust in the system, rated on a 0–10 scale.}
    \label{fig:survey_distributions}
\end{figure}

For assessing the \textit{feature utility} dimension, experts rated the
perceived usefulness of each individual component of the pipeline on a
1--5 Likert scale (Figure~\ref{fig:survey_distributions}a). The
highest-rated components are \emph{KG Citations} ($M = 4.33$,
$SD = 1.41$) and \emph{Storyline \& Fields} ($M = 4.00$, $SD = 1.32$),
indicating that experts find the most value in the structured narrative
summaries and in the source-grounded textual descriptions attached to
knowledge graph elements. \emph{Interactive KG Queries} receive a
moderately positive rating ($M = 3.75$, $SD = 1.75$, $N = 8$), though
the high standard deviation signals polarised opinions---some experts
appreciate the exploratory capability while others find it less
intuitive. Notably, \emph{Causal Graphs} receive the lowest rating
($M = 2.78$, $SD = 1.09$), suggesting that the automatically extracted
causal structures are not yet perceived as sufficiently reliable or
actionable by domain practitioners. This is consistent with Ronco et
al.~\cite{ronco_disaster_2026}, where the usefulness of knowledge
graphs was similarly rated as only ``somewhat useful'' by most
evaluators. It is precisely this limitation that motivated the design
choice, introduced here, of enriching each KG node and link with
automatically generated textual descriptions grounded in source
citations (Section~\ref{sec:factualityKG}). The effectiveness of this
strategy is reflected in the markedly higher ratings received by
\emph{KG Citations}, the highest-scored component in the evaluation,
suggesting that anchoring graph elements to retrievable, source-backed
explanations substantially increases their perceived value and
compensates for the interpretability limitations of the bare graph
structure.

For the \textit{operational impact} dimension, experts rated the system's
potential contribution to humanitarian workflows, providing moderately
positive scores across all three features. 
\emph{Efficiency} and \emph{Situational Awareness} both receive a mean of $3.67$ ($SD = 1.22$ and $1.00$ respectively), indicating that experts see tangible potential for the system to accelerate information synthesis and support a more comprehensive understanding of evolving
crises. \emph{Future Preparation} is rated slightly lower ($M = 3.33$,
$SD = 1.50$), with the higher variance reflecting uncertainty about
whether the system's outputs---derived primarily from past and ongoing
events---can effectively inform preparedness for future disasters. The
consistent placement of all three features above the scale midpoint
is encouraging, though the moderate absolute values suggest that the
system is viewed as a promising complement to, rather than a
replacement for, existing analytical workflows.

Finally, experts assessed \textit{overall trust} in the system, yielding a mean of 6.56 (SD=2.06) on a 0--10 scale (Figure~\ref{fig:survey_distributions}c), with all individual ratings falling within the 4--9 range. 
While these scores indicate adequate rather than high trust, they are
encouraging for a fully automated pipeline requiring no human
intervention and it is coherent with the positive trends reported across preceding evaluation dimensions. The spread of ratings reflects diverse expert
expectations: residual reservations concentrate on causal graph
quality and factual inconsistencies in storyline generation.
The lowest score (4/10) is particularly instructive --- the evaluator
independently traced citations back to their original sources and
found discrepancies between the reported information and the cited
documents. Although such errors were infrequent, this case
illustrates how even isolated citation inaccuracies can
disproportionately erode trust in high-stakes humanitarian contexts,
where every claim is expected to be verifiable.

\section{Discussion and Conclusion}\label{sec:conclusions}
We presented an end-to-end pipeline for generating source-grounded disaster storylines and causal knowledge graphs from heterogeneous humanitarian sources, integrating ReliefWeb as a complementary evidence base, a Multi-Shot RAG strategy with full source traceability, a citation-grounded validation layer for every KG element, and child-sensitive impact dimensions.
Evaluation across three diverse crisis use cases involving 18 independent annotators confirms the pipeline's effectiveness: 85.8\% retrieval precision, 86.7\% of causal triples grounded in source material, and an overall expert trust of 6.56 out of 10. Citation-grounded components received the highest utility ratings ($M{=}4.33$), validating the design choice of anchoring graph elements to retrievable source evidence. Experts highlighted the tool's potential for rapid situational overview, cross-agency comparison of reported figures, and preliminary impact assessment.
The evaluation also surfaces clear limitations, which map onto our future work. Causal graphs received the lowest utility rating ($M{=}2.78$), perceived as oversimplified and potentially misleading without human validation, calling for refined graph complexity and confidence ranges flagging inter-source disagreement. The absence of temporal provenance was identified as a critical gap — storylines lack timestamps showing when figures were reported and how they evolved — motivating timestamped provenance with cross-source reconciliation. Experts further asked for sector-aligned structuring consistent with the humanitarian cluster system and for ingesting user-supplied or restricted-access documents.

Finally, storyline citations proved markedly less reliable than KG ones
($71.7$\% vs.\ above $92$\%), as the model keeps citing even when it lacks
the evidence to answer --- a behaviour that explicit abstention could mitigate.

On the operational side, we plan to run the pipeline over the full EM-DAT catalogue and publicly release the resulting dataset: a source-grounded (EMM and ReliefWeb), narrative-enriched version of EM-DAT offering contextual detail beyond the aggregate statistics currently available per event.

\begin{acks}
The European Union owns the copyright of this work. © European Union, 2026.

The authors acknowledge support from the Lagrange Project of the ISI Foundation, funded by Fondazione CRT
\end{acks}

\bibliographystyle{ACM-Reference-Format}
\bibliography{bibliography}


\end{document}